# Finding Hidden Relationships Between Medical Concepts by Leveraging Metamap and Text Mining Techniques

Weikang Yang[1], S M Mazharul Hoque Chowdhury[2], and Wei Jin[3]

[1]Department of Computer Science, North Dakota State University, Fargo, USA (yangvigoo@gmail.com)

[2]Department of Computer Science and Engineering, Daffodil International University, Dhaka, Bangladesh (mazharul2213@diu.edu.bd)

[3]Department of Computer Science and Engineering, University of North Texas, Denton, USA (wei.jin@unt.edu)

**Abstract—**Text is one of the most common ways to store data in this computerized world. At a glance, it may seem that those data are not interconnected. But in reality, data can have hidden connections. Therefore, in this research, a new model has been presented that can find hidden relationships between two medical concepts by using MetaMap and appropriate text-mining techniques. Specifically, the model creates a new comprehensive index structure and can find cross-document hidden links connecting topics of interest that most existing approaches have ignored. Experiments show the effectiveness of the proposed model in discovering new connections between topics.

## I. INTRODUCTION

If we consider the modern world, it is not possible to imagine it without the use of text. Therefore, text data plays a very important role in every sector of our lives, and most of the time a piece of text type data may contain some connection with other texts. In some cases, those connections are clearly visible and, in some cases, relations remain hidden. On the other hand, when people are interacting with text while reading books or newspapers or communicating with others via text messages, they try to make some connections between different things and try to get hidden information based on the text [1]. However, it is difficult for people to find all the connections based on a little information only. In some cases, a sentence indicating "A implies B" in one article and another sentence "B implies C" in another article, and people generally could not link them together to establish the relationship that "A may imply C", especially when they are facing a large volume of data. This idea was explained by Swanson in 1999 and he proposed a model named 'ABC model' to solve it [2].

The primary focus of this research is motivated by Swanson's ABC model, and we extend it to a cross-document discovery setting and perform a finer granularity of relationship discovery. Additionally, to cope with a large-scale document collection, we introduce a new and efficient index structure that can support multi-level knowledge discovery and we believe it will also facilitate many other downstream natural language processing applications. We further develop a new approach that will be able to predict hidden relations between two input topics effectively. This model was tested against other existing models and showed its effectiveness. Moreover, instead of a single-thread implementation, we also explored a multi-thread implementation setting in order to reduce analysis time and increase system efficiency at the same time. Response time and accuracy can both be further improved by adding threads if computer hardware resource is concerned. In this research Medline database was used as the data source because of its large volume of articles on Bio-medicine and MetaMap was used to analyze the data. Output produced from this model will be combined with Swanson's model to find hidden connections.

The reminder of this paper is structured as follows. Section II discusses related work. In Section III, we present an overview of our methods. In Section IV, we present details of our proposed techniques. Sections V and VI present experiments and evaluation results. And finally, Section VII brings conclusion and gives directions for future work.

## II. LITERATURE REVIEW

In this modern world, data is everything and most of them are text-type data. There are a lot of tools available for text data analysis and researchers around the world have developed tools like MicroMeSH, SAPHIRE, and MetaMap to do this job. MetaMap is a public tool developed by the National Library of Medicine (NLM), which can map text to biomedical concepts using its enormous thesaurus [3]. This has been a popular tool applied to many Information Retrieval and text-mining applications [4]. When MetaMap receives a sentence, it breaks a sentence into phrases and those phrases are used to find mapping concepts to calculate a mapping score for each concept. There are already many applications of MetaMap in the field of biomedical studies. For example, Wendy Marcelo et al. developed a system to detect patients with respiratory illness by processing patients' clinical reports with MetaMap [5]. Zuccon Holloway et al. tried to automatically identify disorders mentioned in health records such as discharge summaries by using MetaMap [6]. In order to test the performance of MetaMap, Pratt and Yetisgen conduct research, which aims to compare MetaMap's capability with that of people who are familiar with the biomedical field [7]. The result shows that MetaMap is capable to identify the biomedical concepts with a 93.3% recall compared with those tagged by biomedical experts.

Knowledge discovery in biomedical texts originates from Swanson's ABC model, based on which Jin and Srihari proposed a new type of query, namely, concept chain queries, attempting to detect hidden links between concepts [2]. Researchers Wei Jin and Rohini Srihari tried to generate concept chains connecting topics of interest from counterterrorism documents. The approach used a variant of the Term Frequency (TF) - Inverse Document Frequency (IDF) weighting scheme for evaluating each potential connecting term, achieving an approximate 82.5% recall [8]. This work is different from Jin's study by using a completely different context – the biomedical domain instead of the counterterrorism corpus. Another related work was proposed by Gopalakrishnan, Kishlay et al., which introduced a new approach that created a graph-based knowledge base and then used a bi-directional search for finding paths in the graph to answer concept chain queries in the biomedical field [9].

Researcher Vishrawas Gopalakrishnan et al. worked on Hypothesis generation and there he used medical data in order to find hidden relations [16]. In their research, they also focused on the hidden relation discovery based on given keywords to develop a hypothesis on a topic. Researcher Xiaohua Hu et al. presented a model that is capable to find hidden links from Medline data and generate novel hypothesis between these concepts [17]. Padmini Srinivasan and Bisharah Libbus worked on research where they used Medline citations to find connections between dietary substances and diseases [18]. In their research, they were trying to discover the therapeutic potential of curcumin/turmeric which is commonly used in Asia. Another research was done by Kishlay Jha and Wei Jin on the extraction of novel knowledge from biomedical literature and domain knowledge [19]. In their research, they presented a meticulous analysis of how manifold statistical information measures and semantic knowledge affect the knowledge discovery procedure.

Moreover, the use of hidden relation extraction covers a large area that includes relations between flood and hydrological variables, immunological markers in multiple sclerosis, and so on. In the year 2007, M.J. Sanscartier and E. Neufeld worked on a model that will be able to find hidden relationships between different variables in a dataset in order to correct models by finding hidden contextual variables [11]. Prakash & Surendran (2013) worked on the detection of social media hidden activity based on the connection between different types of social media profiles and the network traffic of a user [12]. Milton Pividori et al. worked on the discovery of hidden relations between different qualitative and quantitative data without standardization based on their influence on the cluster using a biological dataset [13]. In 2021, Mohamad Basel Al Sawaf et al. worked on a recurrent neural network model that will be able to analyze relationships between different hydrological variables related to flood in order to predict flooding [14]. Amir Hossein Rasekh et al. presented a model that can create descriptions of programming codes that have a relation to associated text documents [15]. Those descriptions can help a user to understand the program logically with little effort.

## III. METHODOLOGY

This research applied the principle of pipe and filter software architectural design [10]. As like in this architecture the original data flow through different modules or filters and in each filter data gets modified and passed through the next one. Finally, the system produces the desired output at the end of the flow.

At first two topics will be provided to the system along with a number of preprocessed documents in order to discover hidden relations. The system will look for relations using the algorithm that can express both topics. Finally, all the findings will be presented as output. As an example, consider the task stated in [20], the first task that pioneered LBD in the biological domain - determining the relationship between fish oil and Raynaud's disease and how they are related.

The system contains three main phases – The MetaMap processing phase (will process and extract documents), the preparation phase (will generate S2C – Source to Concept and fetch the title and abstract), and the output phase (will find relations using Swanson's ABC model). Findings from the output module will be presented using a Graphical User Interface (GUI). This will show the generated concept chains and relevant evidence. Figure 1 presents the data flow of the system.

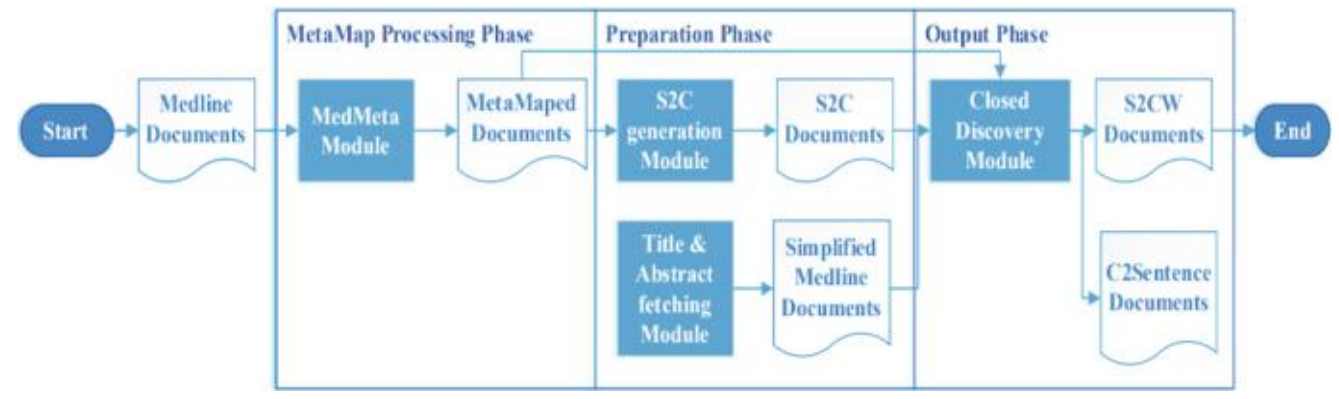

Fig. 1. Data flow diagram

## IV. DATA COLLECTION AND ANALYSIS

The Medline database contains a very large number of biomedical data-related citations and those were released by NLM on their official website. In this research, altogether nearly 22,380,000 citations were used to find hidden relations between two different topics. Here, each citation is an individual document. The Medline database includes a number of attributes such as ID, Publishing date, title, and abstract. For the Hidden Relationship Discovery title and abstract was used to find common links among topics.

### A. Data Extraction from the Source

Data were extracted from the following link: ftp://ftp.ncbi.nlm.nih.gov/pubmed/baseline and XML tags as <PMID>, <PubDate>, <ArticleTitle>, and <AbstractText> were used to identify useful information.

### B. MetaMap Module – Processing Phase

The functional structure of the MedMeta module is divided into four major steps:

a. Extracts useful information from the data source – Medline documents.

b. Inserts MetaMap API in the code and sends titles and abstracts to the MetaMap Server.

c. Reads from MetaMap output and builds indexes of Medline documents based on concept occurrence relationship.

d. Write index files in XML format.

The MedMeta Module consists of seven modules – MedMeta, XMLParser, MedlineCite, MetaMapProc, PostProc, OutputXML, and Thread. The sequence is presented in Fig. 2.

MedMeta's primary object is to regulate the creation of other modules and XMLParser can process oversized XML files. In this research total data used was 93.7 GB in total, where a single file was around 200 MB. After the parsing, XMLParser stored data into MedlineCite (a placeholder module to store data) objects. MetaMapProc will communicate with server and build index to initiate post-processing in PostProc. The PostProc is divided into 3 levels of data structure. The first level is Semantic type (holds one or many Concepts objects), the second one is Concept type (holds one or many Occurrence objects) and the third one is Occurrence type (holds the entity information). OutputXML uses an XML file that will hold the data. The thread module will apply parallel processing to improve performance.

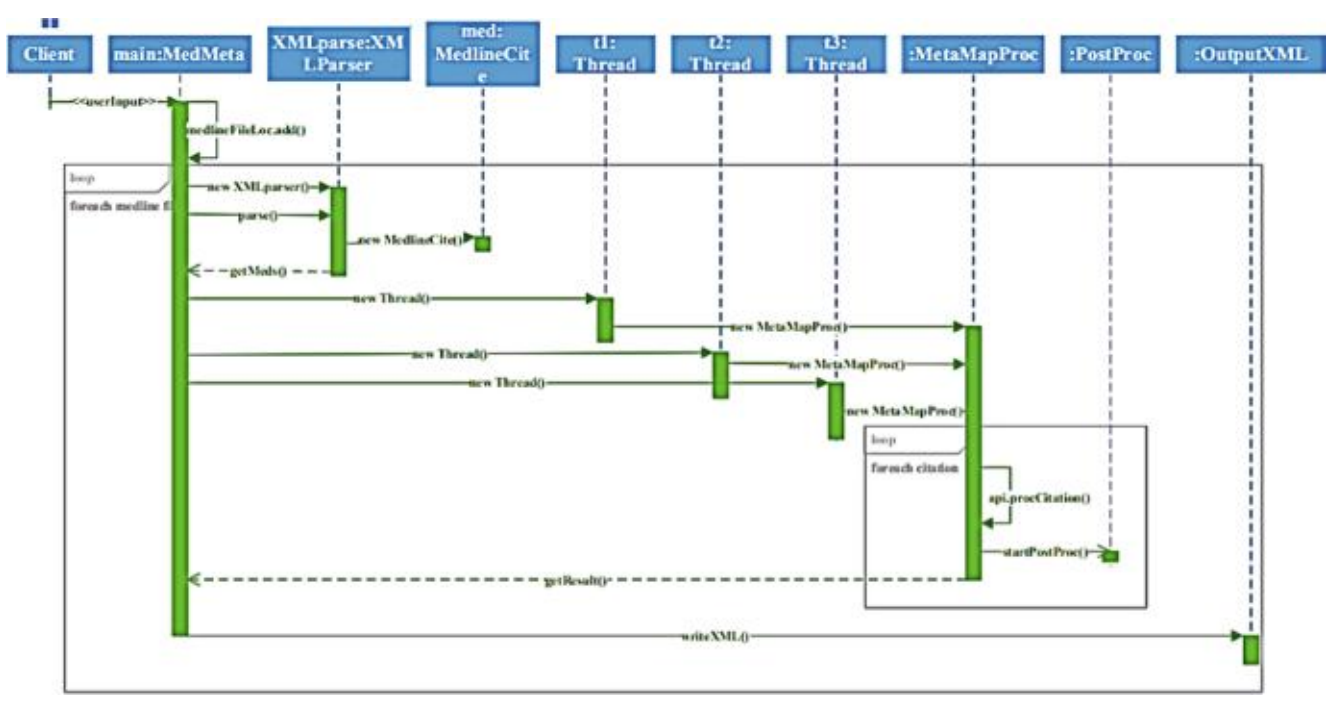

Fig. 2. Sequence diagram for MedMeta module

## C. MetaMap Module – Preparation Phase

In the preparation phase, a simplified S2C document will be produced using semantic and concept type data structure to reduce access time complexity. During the data collection phase, OutputXML generated a big amount of data and those were converted to 2 level S2C to reduce size. According to Fig. 3, S2C generation module consists of four modules – S2CGenerator (forwards the document through XMLParser), XMLParser (storing data in the server), SemanticType (uniquely leveling each hash to hold the S2C relationship), and WriteXMLFile (writes relationships in the XML file). Here, S2CGenerator is responsible for forwarding the document through XMLParser. There are 133 semantic types in total and each of them holds a hash set of concepts.

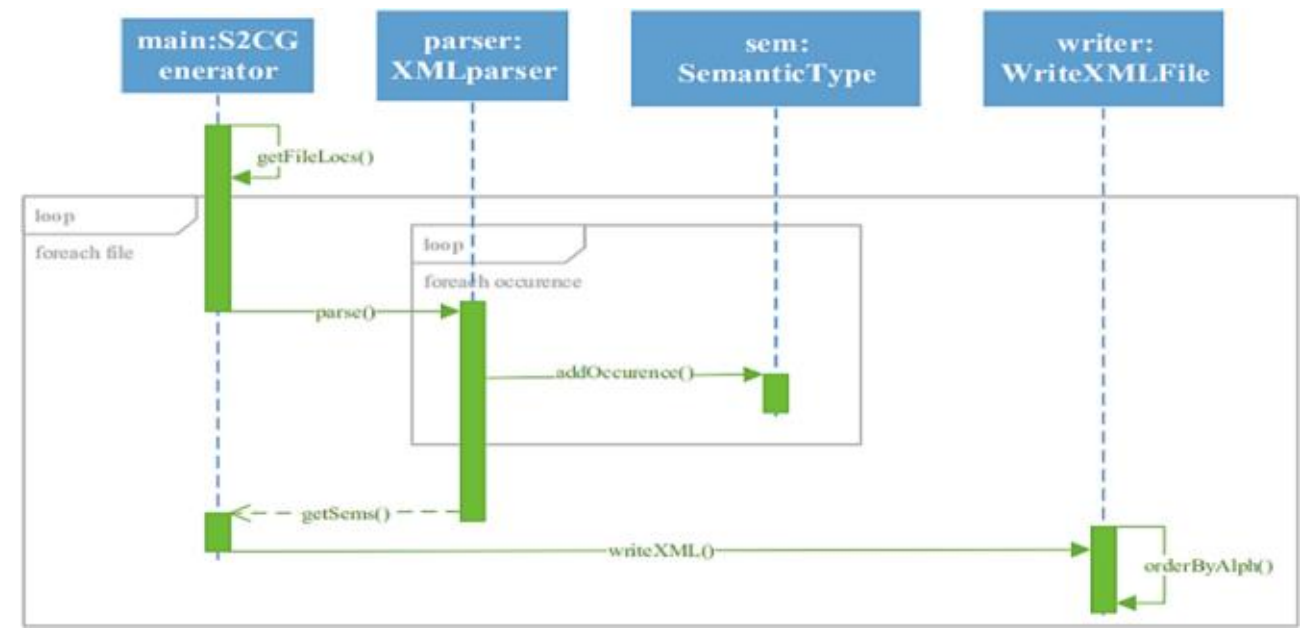

Fig. 3. Sequence diagram for S2C generation module

## D. Title and Abstract Fetching Module

Aim of this module is to fetch the title and abstract from the original document in order to reduce the access time complexity. Figure 4 shows the title and abstract fetching module.

The Title and Abstract fetching module consist of four modules. Three of them are TNAFetcher, XMLParser, and WriteXMLFile and their design are the same as their counterpart in the S2C generation module. Besides, the fourth one is MedLineCitation and it has two data fields – Title and Abstract, where data fetched from XML documents gets stored. Figure 5 shows a simplified Medline Document.

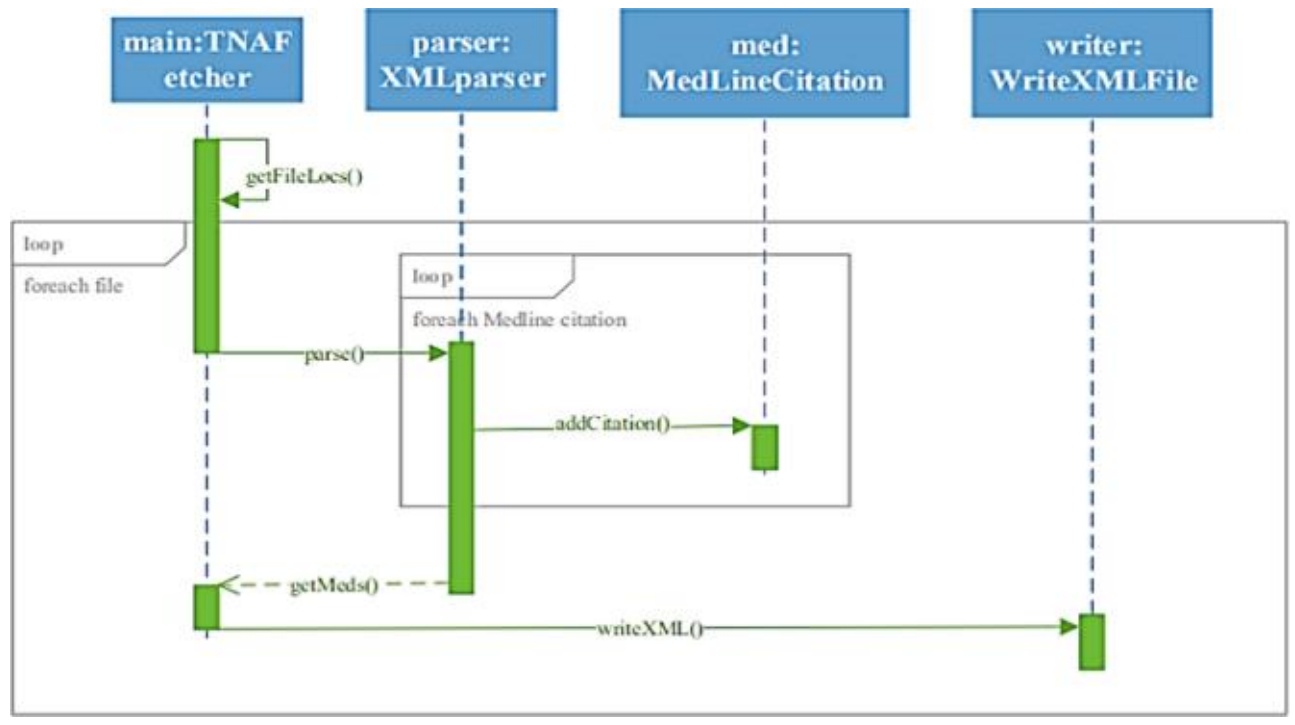

Fig. 4. Sequence diagram for title & abstract fetching module

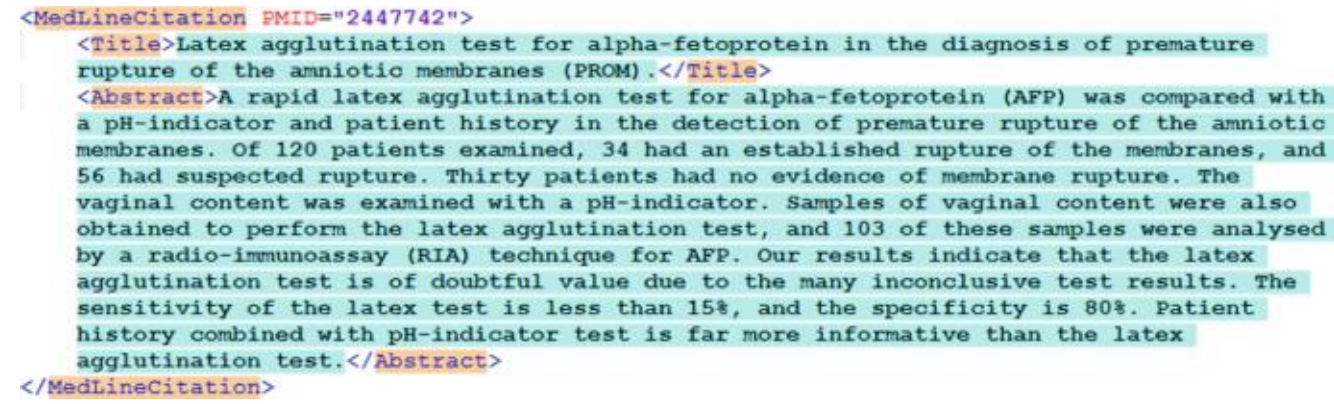

```
<MedLineCitation PMID="2447742">
    <Title>Latex agglutination test for alpha-fetoprotein in the diagnosis of premature
    rupture of the amniotic membranes (PROM).</Title>
    <Abstract>A rapid latex agglutination test for alpha-fetoprotein (AFP) was compared with
    a pH-indicator and patient history in the detection of premature rupture of the amniotic
    membranes. Of 120 patients examined, 34 had an established rupture of the membranes, and
    56 had suspected rupture. Thirty patients had no evidence of membrane rupture. The
    vaginal content was examined with a pH-indicator. Samples of vaginal content were also
    obtained to perform the latex agglutination test, and 103 of these samples were analysed
    by a radio-immunoassay (RIA) technique for AFP. Our results indicate that the latex
    agglutination test is of doubtful value due to the many inconclusive test results. The
    sensitivity of the latex test is less than 15%, and the specificity is 80%. Patient
    history combined with pH-indicator test is far more informative than the latex
    agglutination test.</Abstract>
</MedLineCitation>
```

Fig. 5. A simplified medline document

## E. Closed Discovery Module

This module uses user inputs, MetaMaped documents, S2C documents, and simplified Medline documents as inputs and after processing generates S2CW files and displays outputs through a Graphical User Interface. Figure 8 shows the Sequence diagram of the Closed Discovery module (Fig. 6).

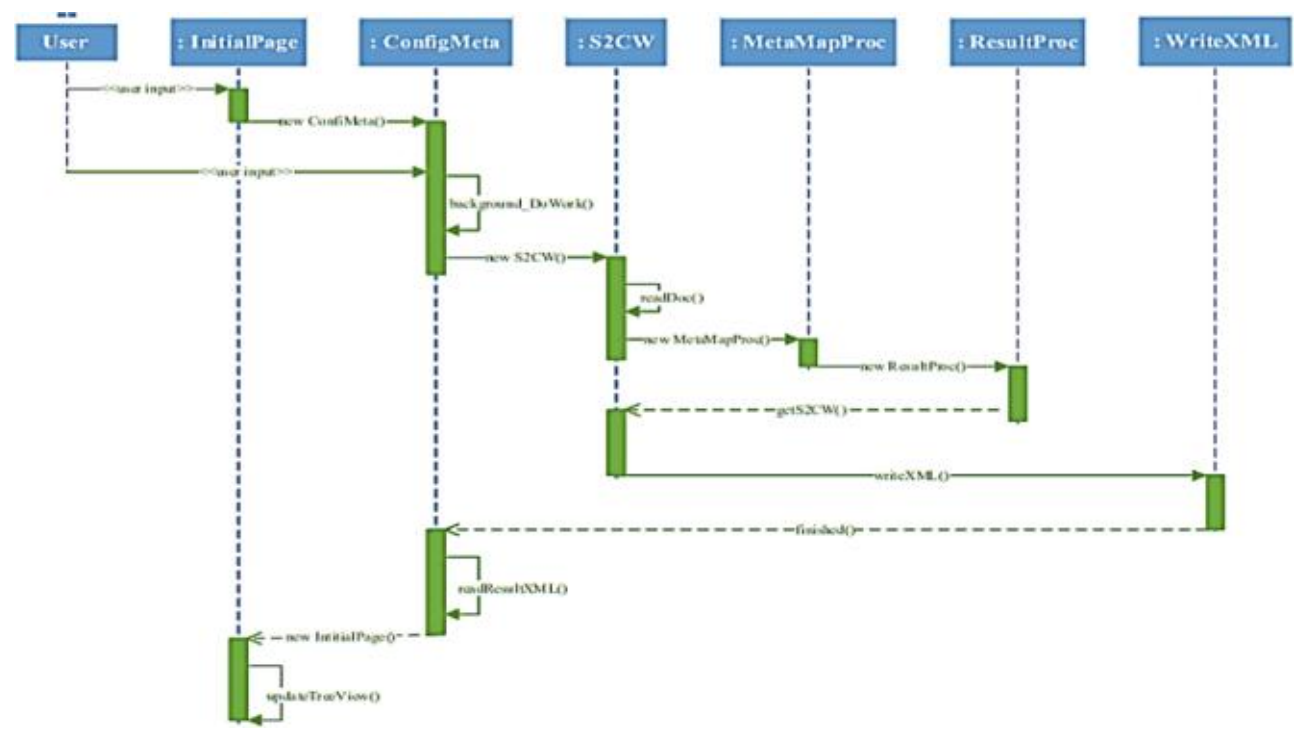

Fig. 6. Sequence diagram of closed discovery module

The ClosedDiscovery Module consists of eight modules – InitialPage, ConfigMeta, S2CW, MetaMapProc, ResultProc, WriteXML, SentPage, and MergedSentPage. SentPage and MergedSentPage are not included in Fig. 6, but they are used to display additional graphical user interfaces. Here, the InitialPage of GUI takes user inputs "Topic 1" and "Topic 2" and through ConfigMeta user can select target data. ConfigMeta is also responsible for building XML documents to store sentences related to Topic 1 and Topic 2. S2CW passes the XML document to MetaMapProc, which works as a gateway to the server that return findings. ResultProc uses those findings to build concept chains by calculating Term Frequency (TF) and Inverse Document Frequency (IDF). The IDF measures the importance of the concept and is measured for concept c by:

$$IDF(c) = log_e(Total\ number\ of\ sentences\ /\ Number\ of\ sentences\ with\ concept\ c\ in\ it) \quad (1)$$

The assumption is that the rarer a concept appears in the sentences, the higher the IDF value is, meaning the more important the concept is to the context. Then, the ResultProc multiplies TF and IDF to get concept weights (i.e. $Weight_e = TF_e * IDF_e$). Each MetaMap may contain several concepts and ResultProc analyze them to build C2W (Concept to Weight) and C2Sents (Concept to Sentence) relationship. Together they create S2CW (Semantic to Concept Weight) relationship, based on which the weight of each concept is further normalized by:

$$Normalized\ Weight(c) = (weight\ of\ concept\ c)\ /\ (maximum\ weight\ in\ this\ Semantic\ Type) \quad (2)$$

After the normalization data will be arranged in the descending order and by merging S2CW for topic 1 (S2CWA) and S2CW for topic 2 (S2CWC) intermediate level linking (common things between Topic 1 and 2) S2CWB will be generated. At this point, WriteXML will build XML files for S2CWA, S2CWB, S2CWC, C2SentsA, and C2SentsC using DocumentBuilderFactory Java API. SentPage is responsible for viewing all sentences related to a certain concept and MergedSentPage will list out all sentences related to concept chains A-B and B-C. Figure 7 represents the GUI which takes the input from the user and visualize the output and Fig. 8 represents linking sentences between topic 1 and 2.

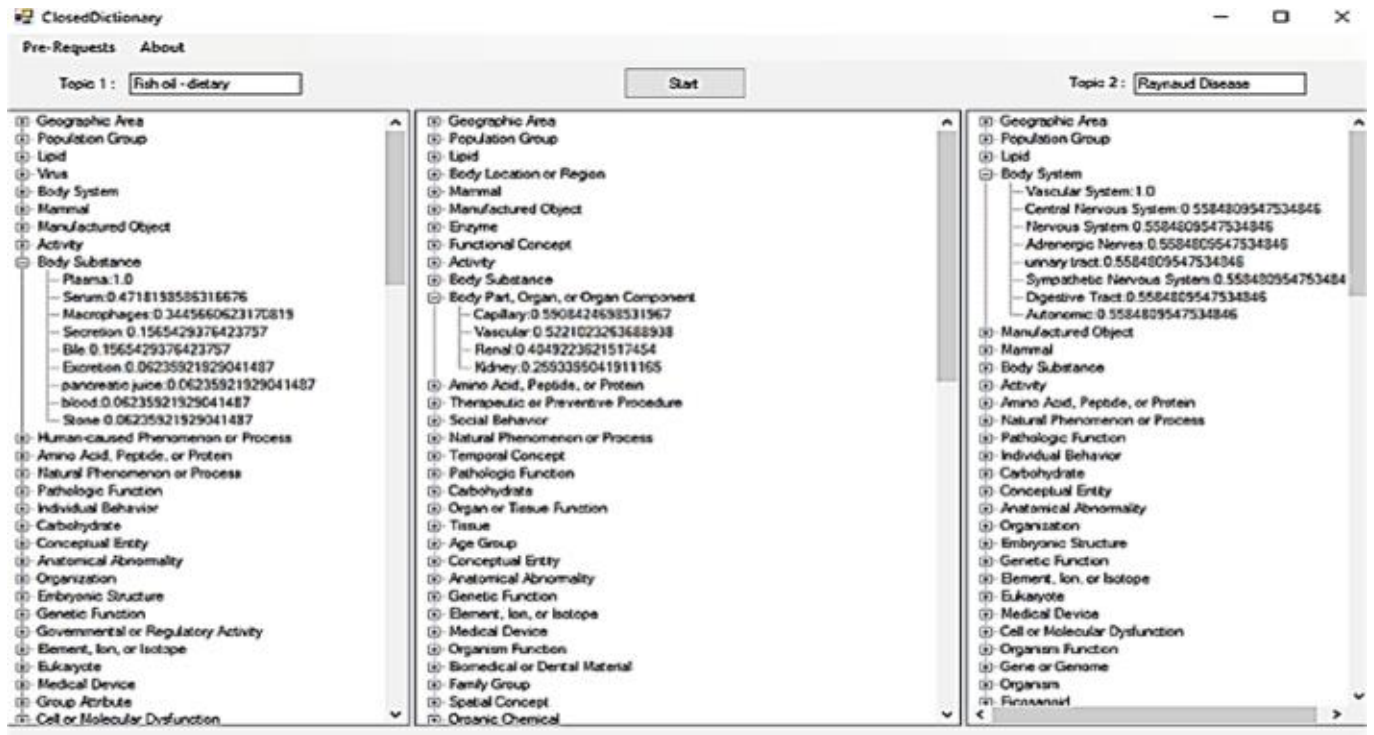

Fig. 7. GUI for visualizing relationship between topics 1 and 2

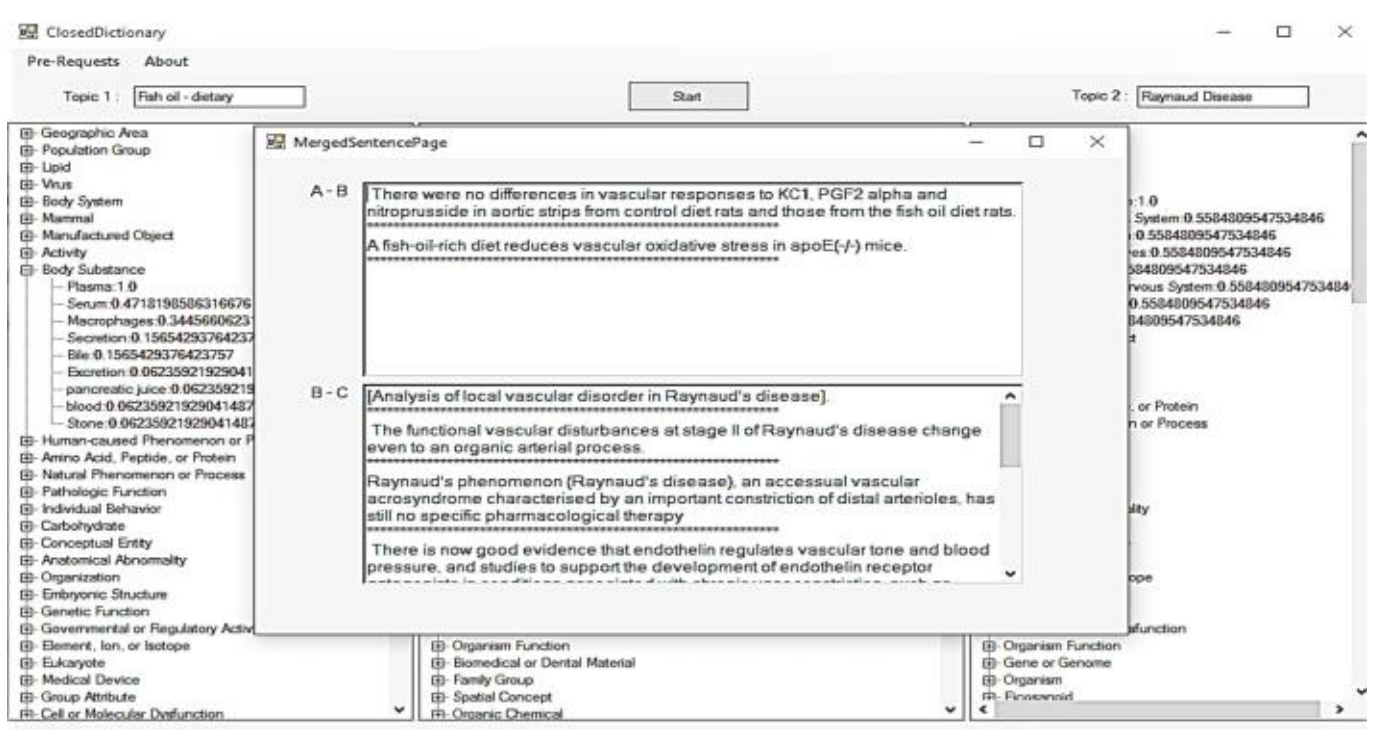

Fig. 8. GUI for sentences that are creating the relationship

## V. SYSTEM EVALUATION

During the development of MedMeta module a multi-thread program (3 threads) was used to communicate with the MetaMap server instead of single thread to evaluate system performance. An experiment was conducted to find the impact of threads against performance with a system that has eight-core i7-4790k CPU, 16GB RAM and Windows Pro 10 x64 OS. The number of threads varies from one to three (Table 1).

According to the table it is clearly visible that with the increase of the number of threads performance increases greatly, where for 1 thread it took 145 s and for 3 threads it took 62 s. Moreover, 3 threads achieve 2.34 times improved performance over a single thread. It shows, in case of hardware limitations three-threads is a preferred option.

TABLE I. PERFORMANCE COMPARISON

| Threads | Proc. Time | Avg Memory | Avg CPU | Perf. vs 1T | Perf. (%) |
|---|---|---|---|---|---|
| 1 | 145 s | 98 MB | 25% | 1 | |
| 2 | 87 s | 110 MB | 40% | 1.67 | 85% |
| 3 | 62 s | 128 MB | 73% | 2.34 | 78% |

## VI. RESULT EVALUATION

In order to justify the accuracy of the model, we used the widely adopted evaluation queries in the related literature, e.g., comparison was made between this research and Gopalakrishnan and Kishlay's study [9] using the gold standard query pair “Fish-oil” and “Raynaud’s Disease”. Gopalakrishnan and Kishlay had found connecting words such as platelet aggregation, vascular reactivity, blood viscosity and Prostaglandin. Our model not only finds all these important connecting terms, but also assigns each a high rank in its associated semantic types. Table 2 presents our partial result of evaluating the above query.

Moreover, our model also found other important linking terms such as “Hemodynamic” and “Atherosclerosis”. However, these links were not detected by them [9].

**TABLE II. EVALUATING THE QUERY PAIR: FISH OIL AND RAYNAUD DISEASE**

| Connecting Concepts | Find? | Weight | Semantic Type Rank |
|---|---|---|---|
| Platelet aggregation | True | 0.87 | 2 |
| Vascular reactivity | True | 1.27 | 1 |
| Prostaglandin | True | 0.309 | 6 |
| Arthritis rheumatoid | True, as "Rheumatoid Arthritis" | 0.43 | 2 |

Another query was made for the pair “Schizophrenia” and “Phospholipase A2”. Gopalakrishnan and Kishlay found linking terms as chlorpromazine, receptors dopamine, prolactin, arachidonic acid, phenothiazines, and norepinephrine [9]. Table 3 shows how our framework behaves in this case. This model was able to find relevant data as them and again this system found interesting linking terms, such as “PGE2”, which [9] could not detect.

**TABLE III. EVALUATING THE QUERY PAIR: SCHIZOPHRENIA AND PHOSPHOLIPASE A2**

| Connecting Concepts | Find? | Weight | Rank in Semantic Type |
|---|---|---|---|
| Chlorpromazine | True | 0.22 | 11 |
| Receptors dopamine | True | 0.33 | 5 |
| Prolactin | True | 1.17 | 2 |
| Arachidonic acid | True | 1.18 | 1 |
| Norepinephrine | True | 0.3 | 11 |

For the standard query pair Migraine and Magnesium, Gopalakrishnan and Kishlay had found connecting words such as propranolol, adenosine triphosphate, calcium, ergotamine, serotonin, norepinephrine, adenine nucleotides, and epinephrine [9]. Table 4 shows how our framework behaves.

Table 4 indicates that the system has also found all those eight connecting concepts found in Gopalakrishnan and Kishlay’s model. Our system also found some other linking terms that were not found in the previous study, such as “Insulin”.

**TABLE IV. EVALUATING THE QUERY PAIR: MIGRAINE AND MAGNESIUM**

| Connecting Concept | Find? | Weight | Rank in Semantic Type |
|---|---|---|---|
| Propranolol | True | 0.41 | 8 |
| Adenosine triphosphate | True | 0.52 | 10 |
| Calcium | True | 1.50 | 1 |
| Ergotamine (as "Ergot Alkaloids") | True | 0.34 | 5 |
| Serotonin | True | 1.49 | 1 |
| Norepinephrine | True | 0.79 | 6 |
| Adenine nucleotides | True | 0.32 | 14 |
| Epinephrine | True | 0.49 | 13 |

## VII. CONCLUSION AND FUTURE WORK

The goal of this research is to find the hidden connections between concepts of interest. This will assist researchers in the early discovery of hypotheses that may lead to important findings by initiating a deep understanding of the hidden information in the biomedical sector. Experiments show that the developed model can find meaningful logical connections between topics of interest and can visualize discovered hypotheses in a user-friendly way. In future work, we will expand the chains to multiple levels, which we call concept graph queries, and MetaMaped files and other additional domain-specific resources will be combined for easier access and more comprehensive knowledge discovery.